\documentclass{article}

\usepackage{wrapfig}
\usepackage{CJKutf8} 
\usepackage{tabularx} 
\usepackage{evocua_style}

\usepackage{xcolor}     
\usepackage{tcolorbox}  

\usepackage{float}
\usepackage{algorithm}
\usepackage{algpseudocode}

\usepackage{pgfplots}
\pgfplotsset{compat=1.18}

\usepackage[noblocks]{authblk}

\makeatletter
\renewcommand\AB@affilsepx{ \quad } 
\makeatother

\usepackage{graphicx}
\usepackage{booktabs}
\usepackage{listings}
\usepackage{xcolor}
\usepackage{subcaption}
\usepackage{geometry}
\definecolor{codegray}{rgb}{0.5,0.5,0.5}
\definecolor{codepurple}{rgb}{0.58,0,0.82}
\definecolor{backcolour}{rgb}{0.95,0.95,0.92}
\lstdefinestyle{jsonstyle}{
    backgroundcolor=\color{backcolour},
    commentstyle=\color{codegray},
    keywordstyle=\color{magenta},
    stringstyle=\color{codepurple},
    basicstyle=\ttfamily\footnotesize,
    breakatwhitespace=false,
    breaklines=true,
    captionpos=b,
    keepspaces=true,
    showspaces=false,
    showstringspaces=false,
    showtabs=false,
    tabsize=2,
    frame=single
}

\usepackage[utf8]{inputenc} 
\usepackage[T1]{fontenc}    
\usepackage{newunicodechar}
\newunicodechar{，}{,}
\usepackage{hyperref}       
\usepackage{xcolor}
\usepackage{svg} 

\hypersetup{
    colorlinks=true,      
    linkcolor=blue,      
    urlcolor=blue,       
    citecolor=blue,      
    linkbordercolor=blue, 
    urlbordercolor=blue,
    citebordercolor=blue,
    pdfborderstyle={/S/U/W 1}, 
}

\usepackage{graphicx}
\usepackage{subcaption}
\usepackage{tcolorbox}
\usepackage{listings}
\usepackage{xcolor}
\usepackage{fontawesome5} 

\definecolor{codegreen}{rgb}{0,0.6,0}
\definecolor{codegray}{rgb}{0.5,0.5,0.5}
\definecolor{codepurple}{rgb}{0.58,0,0.82}
\definecolor{backcolour}{rgb}{0.95,0.95,0.92}

\lstdefinelanguage{json}{
    basicstyle=\ttfamily\footnotesize,
    numbers=none,
    backgroundcolor=\color{backcolour},
    stringstyle=\color{codepurple},
    keywordstyle=\color{blue},
    showstringspaces=false,
    breaklines=true,
    frame=single,
    rulecolor=\color{lightgray},
}

\usepackage{xurl}            
\usepackage{booktabs}       
\usepackage{amsfonts}       
\usepackage{nicefrac}       
\usepackage{microtype}      
\usepackage{lipsum}		
\usepackage{graphicx}
\usepackage{natbib}
\usepackage{doi}
\usepackage{amsmath}
\usepackage{amssymb} 
\usepackage{xspace}
\usepackage{enumitem}
\usepackage{multirow}
\usepackage{multicol}
\usepackage{subcaption} 
\usepackage{makecell}
\usepackage{hyperref, cleveref}
\setlist[itemize]{leftmargin=*}
\setlist[enumerate]{leftmargin=*}
\setlist[description]{leftmargin=*}

\usepackage{colortbl}
\definecolor{mygray}{gray}{.88}
\definecolor{mycyan}{cmyk}{.15,0,0,0}
\definecolor{mycyan2}{cmyk}{.85,0,0,0}

\definecolor{mygreen}{rgb}{0.19, 0.79, 0.02}
\definecolor{midnightgreen}{rgb}{0.0, 0.29, 0.33}

\definecolor{codegray}{rgb}{0.5,0.5,0.5}
\definecolor{codepurple}{rgb}{0.58,0,0.82}
\definecolor{backcolour}{rgb}{0.95,0.95,0.92}

\title{EvoCUA-1.5: Online Reinforcement Learning for Multi-turn Computer-Use Agents}

\author[1,2]{Mianqiu Huang\textsuperscript{*,$\dagger$}}
\author[1]{Taofeng Xue\textsuperscript{*,$\dagger$}}
\author[1]{Chong Peng\textsuperscript{*,$\dagger$}}
\author[1]{Jinrui Ding\textsuperscript{*}}

\author[1,2]{Jie Yang}
\author[1,2]{Sicheng Fan}
\author[1,3]{Jiale Hong}
\author[1,4]{Yufei Gao}
\author[1]{Xiaocheng Zhang}
\author[1]{Linsen Guo}
\author[1]{Xin Yang}
\author[1]{Dengchang Zhao}
\author[1]{Xiandi Ma}
\author[1]{Yuchen Xie}
\author[1]{Peng Pei}
\author[1]{Xunliang Cai}
\author[2]{Xipeng Qiu}

\affil[1]{Meituan}
\affil[2]{Fudan University}
\affil[3]{Shanghai Jiao Tong University}
\affil[4]{Zhejiang University}

\renewcommand{\headeright}{}

\renewcommand{\headeright}{\raisebox{-0.2\height}{\includegraphics[width=5em]{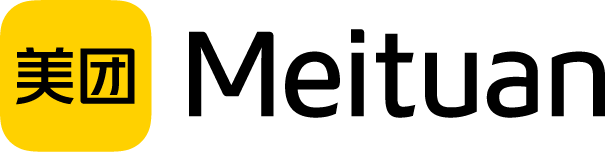}}}

\renewcommand{\shorttitle}{EvoCUA-1.5: Online RL for Computer-Use Agents}

\begin{document}
\maketitle

{
    \renewcommand{\thefootnote}{}
    \footnotetext{
        \textsuperscript{*}Equal contribution. \quad
        \textsuperscript{$\dagger$}Corresponding authors.
    }
}

\vspace{-0.8cm}

\begin{abstract}
Computer-use agents must solve long-horizon tasks through repeated interaction with partially observable, multimodal desktop environments. Although imitation learning and offline trajectory refinement provide strong priors, static traces cannot cover the causal feedback loop of real computer use: each action changes the screen state, future action space, and recovery options. EvoCUA-1.5 extends self-evolving computer-use agents from offline experience learning to online reinforcement learning, where policies interact with executable sandbox environments and improve from verifiable task outcomes.
Online RL in this setting requires more than directly reusing single-turn language-RL recipes. Multi-turn interaction introduces context-managed observations, sparse terminal rewards, variable-length trajectories, and slow environment feedback. EvoCUA-1.5 addresses these challenges with Step-Level Policy Optimization (STEPO), which preserves trajectory-level advantage balance after decomposition into step-level samples; policy-aware filtering and pass-rate calibration over verifiable synthesized tasks; Dynamic Tri-Adaptive Curriculum (DTAC), which combines learnable tasks, difficult positive replay, and controlled infeasible-task exposure; and a fully asynchronous RL infrastructure with staleness control and mini-group batching.
Experiments show that these components improve training stability and downstream performance. EvoCUA-1.5 achieves 63.2\% success on OSWorld-Verified, outperforming comparable 32B/35B-scale open-weight baselines and even approaching models with significantly larger parameter counts. Overall, EvoCUA-1.5 provides a practical framework for scaling online RL in multi-turn computer-use agents.
\end{abstract}

\vspace{-0.6em}
\begin{figure}[H]
    \centering
    \includegraphics[width=0.82\linewidth]{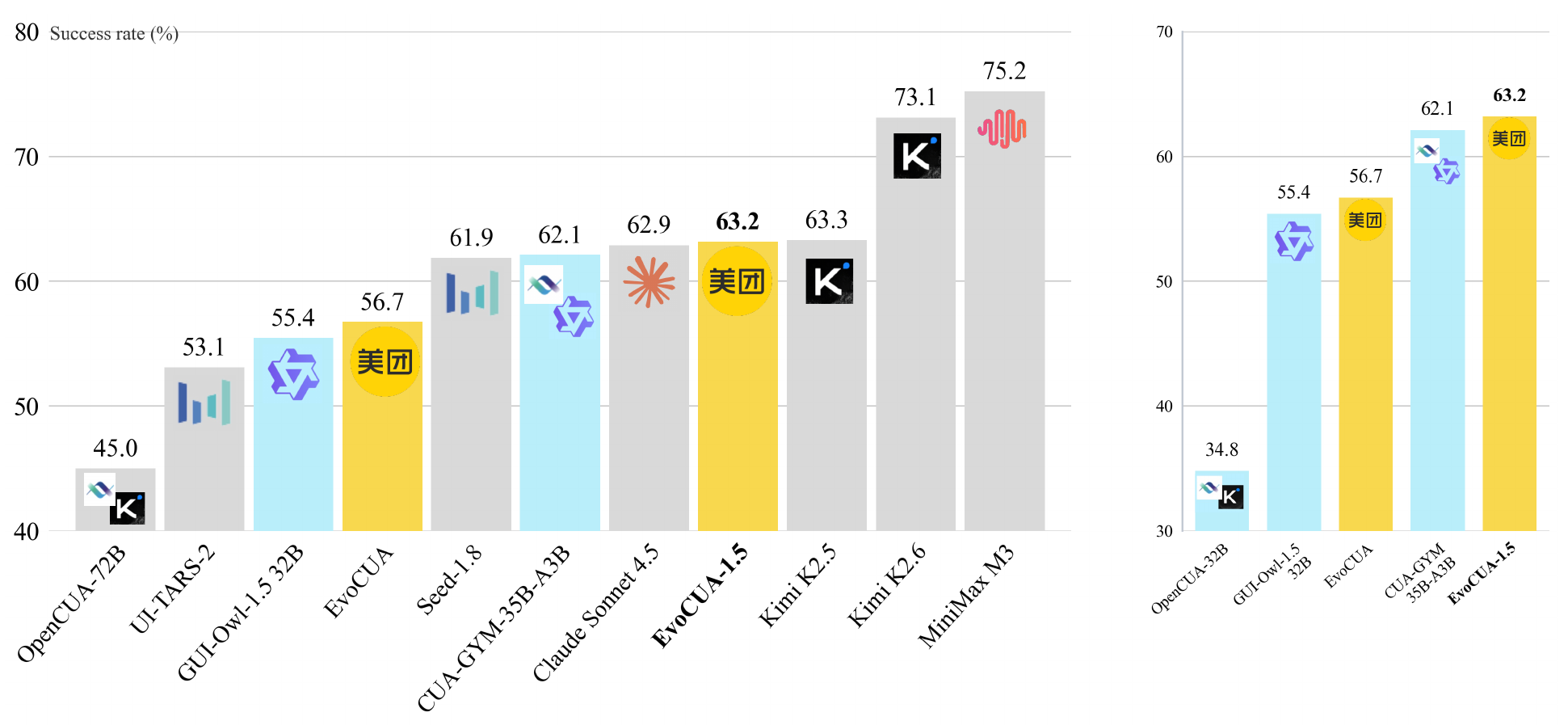}
    \caption{Performance comparison on OSWorld-Verified. EvoCUA-1.5 achieves 63.2\% success rate, outperforming comparable 32B/35B-scale open-weight models.}
    \label{fig:osworld_verified_comparison}
\end{figure}
\vspace{-0.8em}


\newpage

\section{Introduction}

Developing generalist computer-use agents that can operate graphical user interfaces (GUIs) is a critical step toward autonomous computer use. Unlike specialized automation tools, these agents must perceive complex visual contexts, infer task-relevant states from partial observations, and execute long-horizon workflows across heterogeneous applications. Recent native vision-language models (VLMs) have made substantial progress in unifying perception, reasoning, and action within end-to-end computer-use agents~\citep{Qwen3-VL,bytedance2025seed18,wang2025opencua,wang2025ui}. Nevertheless, reliable performance in realistic computer-use environments remains difficult, particularly for tasks that require multi-step exploration, recovery from mistakes, and adaptation to dynamic interface states.

Current progress is still largely driven by imitation learning or offline refinement on fixed interaction traces. Such static data provides useful behavioral priors, but it cannot fully capture the causal feedback loop that characterizes real computer use: each action changes the environment state, which in turn determines the next observation and future action space. Consequently, models trained primarily on offline trajectories are constrained by the coverage of previously collected experience and may struggle with long-tail states, delayed consequences, and failures that emerge only during interaction. This limitation motivates a shift from static trajectory scaling to \emph{online experience scaling}, in which the policy directly interacts with executable environments, receives verifiable feedback, and improves from newly generated experience.

EvoCUA~\citep{xue2026evocua} explored self-evolution from offline synthetic experience. EvoCUA-1.5 extends this line of work by introducing online reinforcement learning for multi-turn computer-use agents. We identify several design choices that make online RL effective in computer-use settings. We find that directly transplanting single-turn RL recipes to computer-use agents is insufficient. A computer-use trajectory contains multiple decisions, and each decision is conditioned on a context-management policy that may evict, fold, or summarize earlier observations. Training samples must therefore be constructed after context management rather than obtained by simply masking a static trajectory. This seemingly implementation-level detail changes the optimization problem: a trajectory-level reward is expanded into multiple step-level samples, and naive GRPO-style advantage reuse can overweight long trajectories and destabilize learning.

Beyond the objective itself, online RL for computer-use agents must also address the efficiency and stability of the full training loop. Because terminal rewards are sparse and trajectories are expensive to generate, poorly calibrated task distributions can waste rollout compute or produce high-variance updates. Tasks that are already too easy provide little contrast, tasks far beyond the current policy mostly yield failed rollouts, and noisy validators can convert environmental feedback into misleading gradients. Meanwhile, the system bottleneck is more severe than in text-only RL: each interaction step involves screenshot rendering, action execution, environment transition, and reward verification, making experience production substantially slower than model optimization. Stable and efficient learning therefore requires coordinated design across objectives, task selection, curriculum sampling, and asynchronous infrastructure.

This paper presents \textbf{EvoCUA-1.5}, an online RL framework for multi-turn computer-use agents. We formalize computer-use interaction with explicit reasoning and context management, and then introduce an integrated set of optimization, data, curriculum, and systems mechanisms for stable online optimization. First, we propose \textbf{Step-Level Policy Optimization (STEPO)}, which computes advantages at the trajectory level and redistributes them across turn-level samples to preserve group-level balance after trajectory decomposition. Second, building on the verifiable task synthesis introduced in EvoCUA, we develop a stricter data filtering and calibration procedure for online RL, using sandbox feasibility checks, executable validators, model-judge trajectory analysis, and policy-dependent pass-rate statistics to maintain high-SNR training data. Third, we propose \textbf{Dynamic Tri-Adaptive Curriculum (DTAC)}, which dynamically combines variance-adaptive sampling, difficulty-adaptive positive replay, and controlled sampling of infeasible tasks. Finally, we build an asynchronous online RL infrastructure that decouples rollout generation from policy updates through a staleness-controlled data buffer.

Our main contributions are summarized as follows:

\begin{itemize}
    \item \textbf{Step-level policy optimization.} We propose \textbf{Step-Level Policy Optimization (STEPO)} to address the mismatch between trajectory-level rewards and turn-level training samples. By distributing each trajectory advantage across its decomposed steps, STEPO improves training efficiency over directly applying GRPO to multi-turn computer-use trajectories.
    \item \textbf{Policy-aware data filtering and calibration.} We introduce a high-SNR online data filtering mechanism built on EvoCUA's verifiable task synthesis, selecting tasks according to sandbox feasibility, validator reliability, and model-dependent pass-rate statistics.
    \item \textbf{Adaptive curriculum learning.} We propose \textbf{Dynamic Tri-Adaptive Curriculum (DTAC)} to dynamically balance learnable tasks, difficult positive replay, and infeasible-task sampling, improving stability under sparse rewards and changing policy capabilities.
    \item \textbf{Asynchronous interaction infrastructure.} We build a staleness-aware asynchronous training infrastructure that decouples experience generation from policy updates, reducing GPU idle time caused by slow environment interaction and variable-length trajectories.
\end{itemize}

Together, these components turn online computer-use interaction into a scalable training signal. Our experiments and ablations show that the resulting framework improves both optimization stability and downstream OSWorld-style performance, while also revealing important practical pitfalls, including model-dependent data selection and potential reward hacking from process reward models.

\section{Preliminaries}

EvoCUA-1.5 inherits the basic computer-use agent formulation from EvoCUA~\citep{xue2026evocua}. We therefore recap only the notation required for online RL and context management. Given a natural-language instruction $g$, the agent interacts with a partially observable computer-use environment. At step $t$, the underlying computer state $s_t$ is rendered as a screenshot observation $o_t=\mathrm{Render}(s_t)$. Conditioned on the retained interaction history $h_t$, the policy generates an explicit reasoning trace $z_t$ and an executable action $a_t$:
\begin{equation}
    \pi_{\theta}(z_t,a_t\mid h_t,o_t).
\end{equation}
After executing $a_t$, the environment transitions to the next state and eventually returns a sparse terminal reward from an executable validator $V_g$:
\begin{equation}
    R(s_T;g) \triangleq \mathbb{I}[V_g(s_T)=\mathrm{True}].
\end{equation}
The key distinction in EvoCUA-1.5 is not the base computer-use interaction formulation, but how online RL should construct training samples and assign trajectory-level rewards under multi-turn context management.

\subsection{Context Management in Multi-Turn Computer-Use Agents}

\begin{figure}[t]
    \centering
    \includegraphics[width=0.95\linewidth]{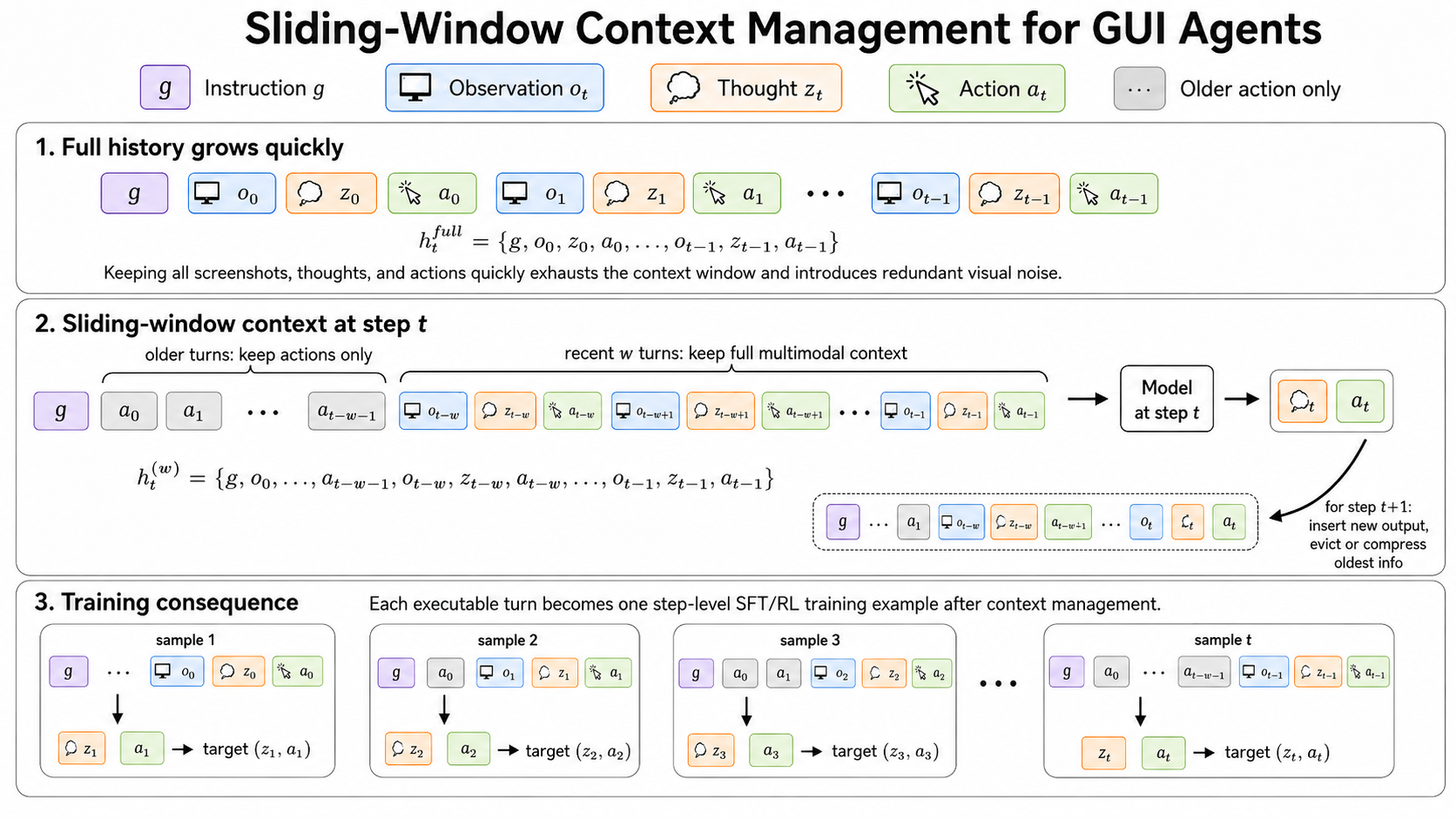}
    \caption{Context management in multi-turn computer-use agent training. The full trajectory history is transformed into a managed context before each decision: earlier turns are compressed into lightweight action history, while recent turns retain full multimodal information. Consequently, training data must be constructed at the step level after context management.}
    \label{fig:context_management}
\end{figure}

Before step $t$, the complete interaction history is
\begin{equation}
    h_t^{full}=\{g,o_0,z_0,a_0,\ldots,o_{t-1},z_{t-1},a_{t-1}\}.
\end{equation}
Feeding this full history directly to the model is impractical for long-horizon computer-use tasks, because screenshots and reasoning traces quickly consume the context window and introduce redundant visual information. Modern computer-use agents therefore rely on context-management strategies such as sliding windows, preserved thinking, folding, and summarization~\citep{wang2025opencua, Qwen3-VL, xue2026evocua}. As illustrated in Figure~\ref{fig:context_management}, this work focuses on the common sliding-window setting. Given a window size $w$ and $t>w$, the managed context stores earlier turns as compact action history and retains full observation--thought--action triples only for the most recent turns:
\begin{equation}
    h_t^{(w)}=\{g,\underbrace{a_0,\ldots,a_{t-w-1}}_{\text{compact action history}},o_{t-w},z_{t-w},a_{t-w},\ldots,o_{t-1},z_{t-1},a_{t-1}\}.
\end{equation}

This context transformation has a direct implication for training. After step $t$ is generated, its reasoning and action are no longer merely prediction targets; they become part of the managed input for step $t+1$, while older information may be evicted or compressed. A multi-turn trajectory therefore cannot be converted into a single static training sequence with a fixed input--output mask. To match inference-time behavior, we construct one training example after each context-management operation. Under the sliding-window strategy, every executable turn becomes a separate step-level sample. The remaining question is how to assign a trajectory-level terminal reward to these step-level samples, which motivates the optimization design in the next section.

\subsection{Learning Objective}

For a group of $G$ rollouts sampled from the same instruction or task family, each trajectory $\tau_i$ contains $|T_i|$ executable turns and receives a terminal reward $R_i\in\{0,1\}$. Online RL improves the policy by contrasting successful and failed trajectories produced by the current or recent policy. The central challenge is to assign trajectory-level rewards to step-level samples without distorting group-level advantage normalization or overweighting long trajectories.

\section{Step-Level Policy Optimization}
\label{sec:stepo}

Objective design is a central challenge for online reinforcement learning in computer-use environments. Unlike a single response conditioned on a fixed prompt, a computer-use trajectory consists of a sequence of interdependent decisions, each conditioned on the evolving visual state and on the agent's context-management policy. The reward, however, is typically observed only at the trajectory level after execution. This section introduces Step-Level Policy Optimization (STEPO), a group-based RL objective adapted to this mismatch between trajectory-level feedback and turn-level training samples.

\subsection{Bias Induced by Naive GRPO under Trajectory Decomposition}

Group Relative Policy Optimization (GRPO) is an appealing post-training objective because it obviates a separate critic while normalizing rewards within a rollout group~\citep{shao2024deepseekmath,guo2025deepseek}. Given a group of $G$ trajectories with terminal rewards $R_1,\ldots,R_G$, the trajectory-level advantage is defined as
\begin{equation}
    A_i = \frac{R_i-\mathrm{mean}(\mathbf{R})}{\mathrm{std}(\mathbf{R})+\epsilon},
    \qquad \sum_{i=1}^{G} A_i \approx 0.
\end{equation}
This zero-mean property is defined over trajectories: before decomposition, each sampled trajectory contributes once to the group update. In a single-turn language task, this is consistent with assigning $A_i$ to all tokens in the only response. In contrast, a computer-use trajectory is naturally expanded into step-level training samples,
\begin{equation}
    \tau_i=\{(x_{i,t},y_{i,t})\}_{t=1}^{|T_i|},
\end{equation}
where $x_{i,t}$ denotes the context-managed observation at step $t$, and $y_{i,t}$ denotes the corresponding reasoning trace and executable action. This expansion is necessary because the output at step $t$ becomes part of the input at step $t+1$, while earlier observations may be evicted, summarized, or compacted by the context-management module. Consequently, the policy must be optimized on the post-context-management view available at each decision point, rather than on a single static trajectory sequence.

A naive application of GRPO after decomposition would assign the same trajectory advantage $A_i$ to every decomposed step. For clarity, abstracting away clipping and KL regularization, the resulting update takes the form
\begin{equation}
    \mathcal{J}_{\mathrm{naive}}(\theta)
    =\mathbb{E}\left[\sum_{i=1}^{G}\sum_{t=1}^{|T_i|}
    \frac{1}{|y_{i,t}|}\sum_{k=1}^{|y_{i,t}|}A_i r_{i,t,k}(\theta)\right],
\end{equation}
where $r_{i,t,k}(\theta)$ is the token-level importance sampling ratio. Aggregating the repeated advantage terms over steps shows that the effective coefficient assigned to trajectory $i$ becomes
\begin{equation}
    \sum_{t=1}^{|T_i|} A_i = |T_i|A_i.
\end{equation}
Thus, naive GRPO no longer corresponds to the original trajectory-balanced objective. Instead, it optimizes a length-weighted surrogate in which longer trajectories induce larger-magnitude updates.

Equivalently, the normalized advantages are no longer balanced over the actual training samples. The sample-level mean advantage becomes
\begin{equation}
    \bar{A}_{\mathrm{sample}}
    =\frac{\sum_{i=1}^{G}\sum_{t=1}^{|T_i|}A_i}{\sum_{i=1}^{G}|T_i|}
    =\frac{\sum_{i=1}^{G}|T_i|A_i}{\sum_{i=1}^{G}|T_i|},
\end{equation}
which is generally nonzero even though $\sum_i A_i\approx 0$. This induces a systematic bias whenever trajectory lengths vary within a rollout group. The issue is particularly consequential in computer-use tasks, where trajectory length often correlates with task difficulty, recovery behavior, and failure modes.

\subsection{Step-Level Policy Optimization}

STEPO corrects this distortion by preserving the trajectory-level advantage mass while still optimizing on step-level samples. For each trajectory, we first compute the GRPO-style advantage $A_i$ and then distribute it uniformly across the steps in that trajectory:
\begin{equation}
    \hat{A}_{i,t}^{\mathrm{STEPO}}=\frac{A_i}{|T_i|}
    =\frac{R_i-\mathrm{mean}(\mathbf{R})}{(\mathrm{std}(\mathbf{R})+\epsilon)|T_i|}.
\end{equation}
This assignment conserves the original trajectory-level advantage:
\begin{equation}
    \sum_{t=1}^{|T_i|}\hat{A}_{i,t}^{\mathrm{STEPO}} = A_i.
\end{equation}
Summing over the full rollout group gives
\begin{equation}
    \sum_{i=1}^{G}\sum_{t=1}^{|T_i|}\hat{A}_{i,t}^{\mathrm{STEPO}}=\sum_{i=1}^{G}A_i\approx 0,
\end{equation}
thereby restoring the group-balance property after trajectory decomposition.

For a turn-level output sequence $y_{i,t}$ that contains both the reasoning trace and the executable action, we optimize the following clipped policy-gradient objective:
\begin{equation}
\begin{split}
    \mathcal{J}_{\mathrm{STEPO}}(\theta)=\mathbb{E}\Bigg[\sum_{i=1}^{G}\sum_{t=1}^{|T_i|}\frac{1}{|y_{i,t}|}\sum_{k=1}^{|y_{i,t}|}
    \min\big(r_{i,t,k}(\theta)\hat{A}_{i,t}, \mathrm{clip}(r_{i,t,k}(\theta),1-\epsilon_c,1+\epsilon_c)\hat{A}_{i,t}\big)\Bigg],
\end{split}
\end{equation}
where $r_{i,t,k}(\theta)$ is the token-level probability ratio between the current policy and the rollout policy, and $\epsilon_c$ is the clipping coefficient. As in PPO and GRPO, a KL penalty to a reference policy can be added to further regularize the update~\citep{schulman2017proximal,shao2024deepseekmath}.

\subsection{Length Preference and Practical Implications}

STEPO also introduces a mild length-dependent inductive bias. For successful trajectories, shorter solutions receive larger per-step positive advantages, which encourages concise execution. For failed trajectories, longer attempts receive smaller-magnitude per-step negative advantages, allowing the policy to continue exploration before terminating or giving up. This bias is desirable in our setting because computer-use tasks often benefit from concise successful execution while still requiring tolerance for exploratory recovery under failure. Nevertheless, it should be monitored in environments where repeated actions can manipulate the reward signal or exploit the validator.

\section{Policy-Aware Data Filtering and Curriculum Learning}
\label{sec:data_curriculum}

A second challenge is deciding which interaction data should be used for online RL. For candidate generation, EvoCUA-1.5 follows the verifiable task synthesis pipeline introduced in EvoCUA~\citep{xue2026evocua}: tasks are generated from environment priors and atomic computer-use abilities, and are paired with sandbox configurations and executable validators. Starting from this synthesized task pool, our focus is the downstream data-selection process: filtering unreliable or uninformative tasks, calibrating task difficulty against the current policy, and scheduling tasks into a policy-aware curriculum for online RL.

\subsection{From Synthesized Tasks to RL-Ready Data}

A synthesized task is not necessarily suitable for online RL. Even when both the instruction and validator are executable, the task may be too easy for the current model, too difficult to yield informative positive trajectories, or too noisy due to brittle environment states. We therefore convert synthesized tasks into RL-ready data through three filtering stages.

\paragraph{Verifiability and feasibility check.}
Each task is executed in the sandbox environment using model rollouts. The executable validator provides terminal success labels, while trajectory analysis helps identify task-definition errors. We distinguish genuine policy failures from ambiguous instructions, invalid initial states, overly strict validators, missing software capabilities, and inherently infeasible goals. Tasks with unreliable configurations or validators are revised or removed so that noisy task definitions are not mistaken for ordinary policy failures.

\paragraph{Atomic-ability coverage control.}
The synthesis pipeline associates each task with one or more atomic computer-use abilities, allowing us to track both ability coverage and the amount of data available for each ability. During filtering, we consider not only task-level quality and pass rate, but also whether the retained pool covers the target ability set with sufficient examples per ability. This prevents online RL from over-representing abilities that are easy to synthesize or validate while under-training abilities that have fewer candidate tasks but remain important for general computer use.

\paragraph{Policy-aware pass-rate calibration.}
For each candidate task, we estimate the pass rate of the current policy using grouped rollouts. Tasks whose pass rates are close to $0$ or $1$ are less informative for online RL: near-zero tasks rarely produce useful successful trajectories, whereas near-one tasks are already largely solved. We therefore prioritize tasks in an intermediate, learnable range, where successful and failed trajectories appear within the same rollout group. Because task difficulty is policy-dependent, this calibration is repeated as training progresses.

\begin{figure}[t]
    \centering
    \includegraphics[width=0.6\linewidth]{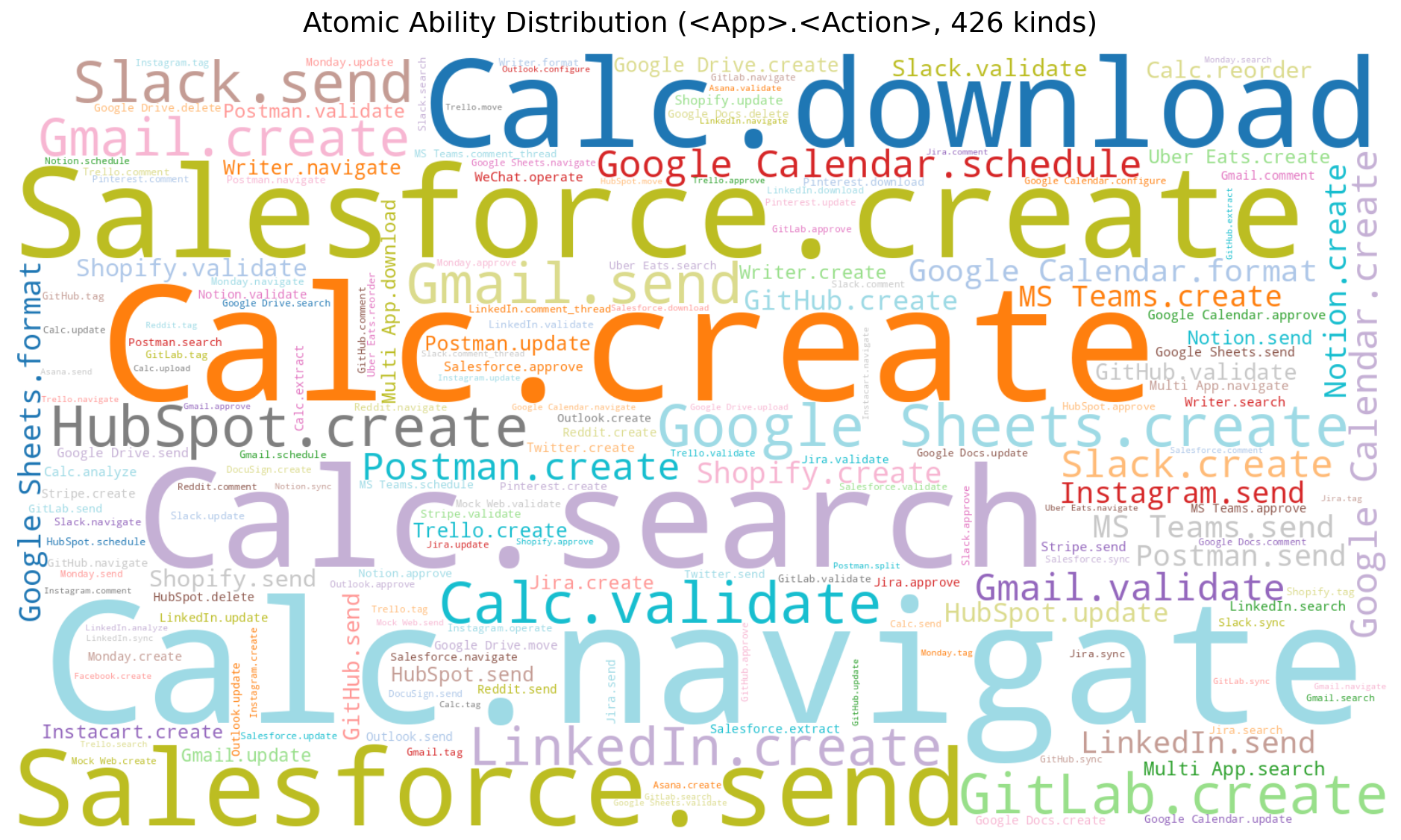}
    \caption{Atomic-ability coverage in the synthesized task pool. EvoCUA-1.5 builds on EvoCUA's verifiable task synthesis pipeline and further filters the task pool through sandbox feasibility checks, validator inspection, pass-rate calibration, and trajectory analysis.}
    \label{fig:task_construction}
\end{figure}

\subsection{Dynamic Tri-Adaptive Curriculum}

The filtering stage produces an RL-ready task pool, but the usefulness of each task changes as the policy improves. A task that is informative early in training may later become trivial, whereas a difficult task may become valuable once the policy begins to solve it occasionally. Static sampling can therefore waste computation on tasks that are either already mastered or not yet learnable. To adapt the data distribution over time, we introduce Dynamic Tri-Adaptive Curriculum (DTAC), which constructs each training batch from three complementary channels:
\begin{equation}
    \mathcal{B}^{(t)}=\mathcal{B}_{\mathrm{VAS}}^{(t)}\cup\mathcal{B}_{\mathrm{AdaPR}}^{(t)}\cup\mathcal{B}_{\mathrm{ICS}}^{(t)}.
\end{equation}
Here, $\mathcal{B}_{\mathrm{VAS}}^{(t)}$, $\mathcal{B}_{\mathrm{AdaPR}}^{(t)}$, and $\mathcal{B}_{\mathrm{ICS}}^{(t)}$ denote the mini-batch components produced by Variance-Adaptive Sampling (VAS), Difficulty-Adaptive Positive Replay (AdaPR), and Infeasibility-Controlled Sampling (ICS), respectively. These channels target complementary regions of the data distribution: currently informative tasks, hard-but-solvable tasks with rare positive trajectories, and infeasible tasks that should be controlled separately.

\begin{figure}[t]
    \centering
    \includegraphics[width=0.95\linewidth]{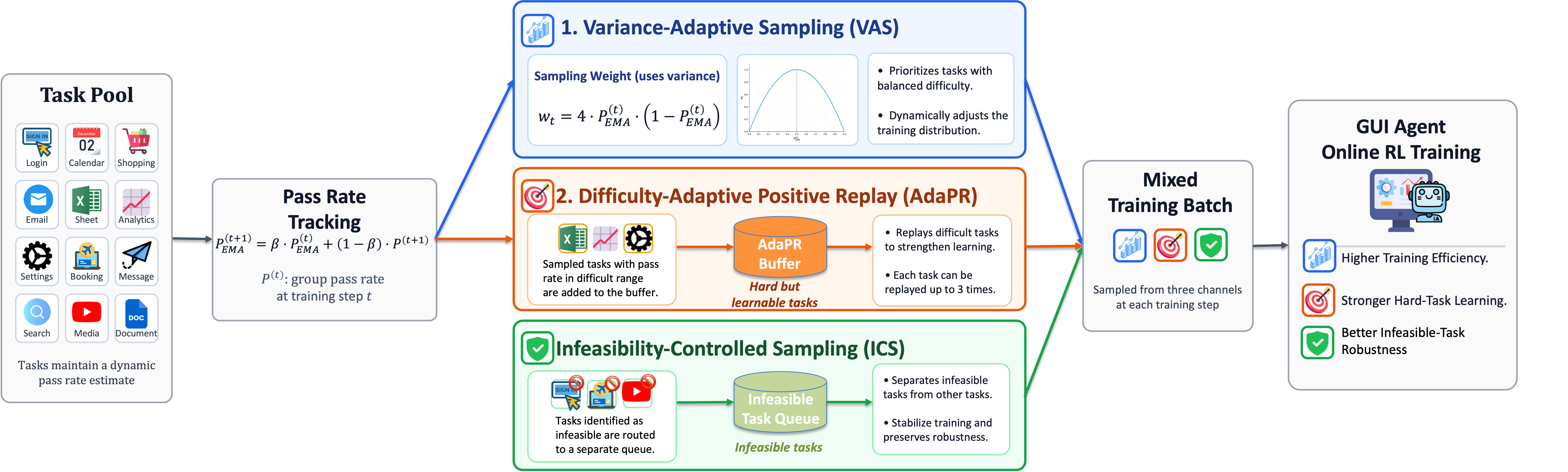}
    \caption{Overview of Dynamic Tri-Adaptive Curriculum (DTAC). Each training batch combines tasks selected by Variance-Adaptive Sampling, hard-but-learnable positive replay examples from AdaPR, and a controlled fraction of infeasible tasks from ICS.}
    \label{fig:dtac_overview}
\end{figure}

Figure~\ref{fig:dtac_overview} illustrates how DTAC combines these channels during curriculum construction.

\subsubsection{Variance-Adaptive Sampling}

The main sampling channel adapts the task distribution according to the current policy's difficulty on each task. In grouped rollouts, tasks with pass rates close to $1$ are already largely solved and provide limited marginal learning signal, whereas tasks with pass rates close to $0$ rarely expose successful trajectories. Tasks with intermediate pass rates are more informative because the same rollout group contains both successes and failures, yielding clearer relative comparisons for policy optimization.

This intuition can be formalized through the uncertainty of a binary success label. If rollout success is modeled as a Bernoulli variable with task pass rate $P$, its entropy is
\begin{equation}
    H(P)=-P\log P-(1-P)\log(1-P),
\end{equation}
which is maximized at $P=0.5$ and decreases as $P$ approaches $0$ or $1$. Thus, medium-difficulty tasks have the highest outcome uncertainty and tend to provide the strongest contrastive learning signal within a rollout group.

Because each training step observes only a limited number of rollouts per task, the raw pass rate can be noisy. We therefore maintain an exponential moving average (EMA) of each task's pass rate:
\begin{equation}
    P_{EMA}^{(t)}=\beta P_{EMA}^{(t-1)}+(1-\beta)P^{(t)},
\end{equation}
where $P^{(t)}$ is the observed group pass rate at training step $t$ and $\beta$ is the smoothing coefficient. The EMA accumulates historical difficulty estimates and prevents the sampling weight from changing abruptly due to accidental successes or failures in a single rollout group.

For efficient sampling, we use the variance of the Bernoulli reward as a normalized proxy for task informativeness. With binary terminal rewards, $r\sim\mathrm{Bern}(P)$ and
\begin{equation}
    \mathrm{Var}(r)=P(1-P).
\end{equation}
This variance has the same qualitative shape as the entropy above: it peaks at $P=0.5$ and vanishes near $0$ or $1$. We therefore define the VAS sampling weight as
\begin{equation}
    w^{(t)}=4P_{EMA}^{(t)}(1-P_{EMA}^{(t)}),
\end{equation}
where the factor $4$ normalizes the weight to $[0,1]$ with maximum value $1$ at $P_{EMA}^{(t)}=0.5$. As training progresses, mastered tasks and persistently unsolved tasks are automatically down-weighted, while tasks that enter a partially solvable regime receive higher sampling probability. VAS therefore avoids fixed difficulty labels and continuously focuses online RL on tasks that are currently most learnable for the policy.

\subsubsection{Difficulty-Adaptive Positive Replay}

VAS assigns low sampling weights to tasks whose pass rates are close to either $0$ or $1$. This is desirable for tasks that are already mastered, but a low pass rate does not necessarily imply that a task is uninformative. For difficult tasks that the policy solves only occasionally, rare successful or partially successful trajectories can provide valuable learning signals. If such tasks rely only on VAS, they may be sampled too infrequently because their overall pass rates remain low, making it difficult for the model to consolidate progress on high-difficulty tasks.

To compensate for this effect, we introduce Difficulty-Adaptive Positive Replay (AdaPR). AdaPR focuses on tasks whose smoothed pass rates fall within a hard-but-learnable interval:
\begin{equation}
    P_{low}\leq P_{EMA}^{(t)}\leq P_{high}.
\end{equation}
Tasks in this interval are challenging for the current policy but not completely out of reach: the model has already produced successful trajectories in some rollout groups, indicating that the task contains actionable positive signal. We add successful or partially successful rollout groups from these tasks to the AdaPR buffer and sample an additional fraction of them in later training steps.

To prevent replay from over-amplifying a small number of hard examples, each buffered item is assigned a maximum replay count. This cap allows the model to revisit rare positive trajectories from difficult tasks while preventing the overall training distribution from being dominated by a few replayed groups. In combination, VAS improves the efficiency of the global sampling distribution, while AdaPR provides additional consolidation for hard-but-solvable tasks that would otherwise be under-sampled.

\subsubsection{Infeasibility-Controlled Sampling}

Finally, infeasible tasks require different treatment from merely difficult tasks. Infeasibility may arise from environment restrictions, invalid goals, missing software capabilities, or intentionally impossible user requests. Treating these tasks as ordinary failures can distort the curriculum, whereas removing them entirely weakens the model's ability to recognize when continued exploration is unproductive. We therefore maintain a separate infeasible-task set $\mathcal{D}_{inf}$ and sample it through an independent channel:
\begin{equation}
    |\mathcal{B}_{\mathrm{ICS}}^{(t)}|=\lfloor \rho_{inf}|\mathcal{B}^{(t)}|\rfloor,
\end{equation}
where $\rho_{inf}$ is a small fixed ratio. This controlled channel exposes the model to infeasible cases at a stable rate, teaching it when to report failure or terminate exploration without allowing infeasible tasks to dominate the main optimization signal.

\section{Asynchronous Online RL Infrastructure}

Large-scale online RL for computer-use agents is bottlenecked primarily by environment interaction rather than by policy optimization alone. Unlike text-only rollouts, a computer-use trajectory requires repeated visual rendering, action execution, environment stabilization, state transition, and validator checks. Each step may take several seconds, and trajectory length varies with task difficulty, application latency, and recovery behavior. A synchronous RL pipeline therefore produces substantial idle time: training GPUs wait for slow environment rollouts, while rollout workers may be blocked by policy updates or unavailable model weights.

\subsection{Architecture Overview}

\begin{figure}[t]
    \centering
    \includegraphics[width=0.95\linewidth]{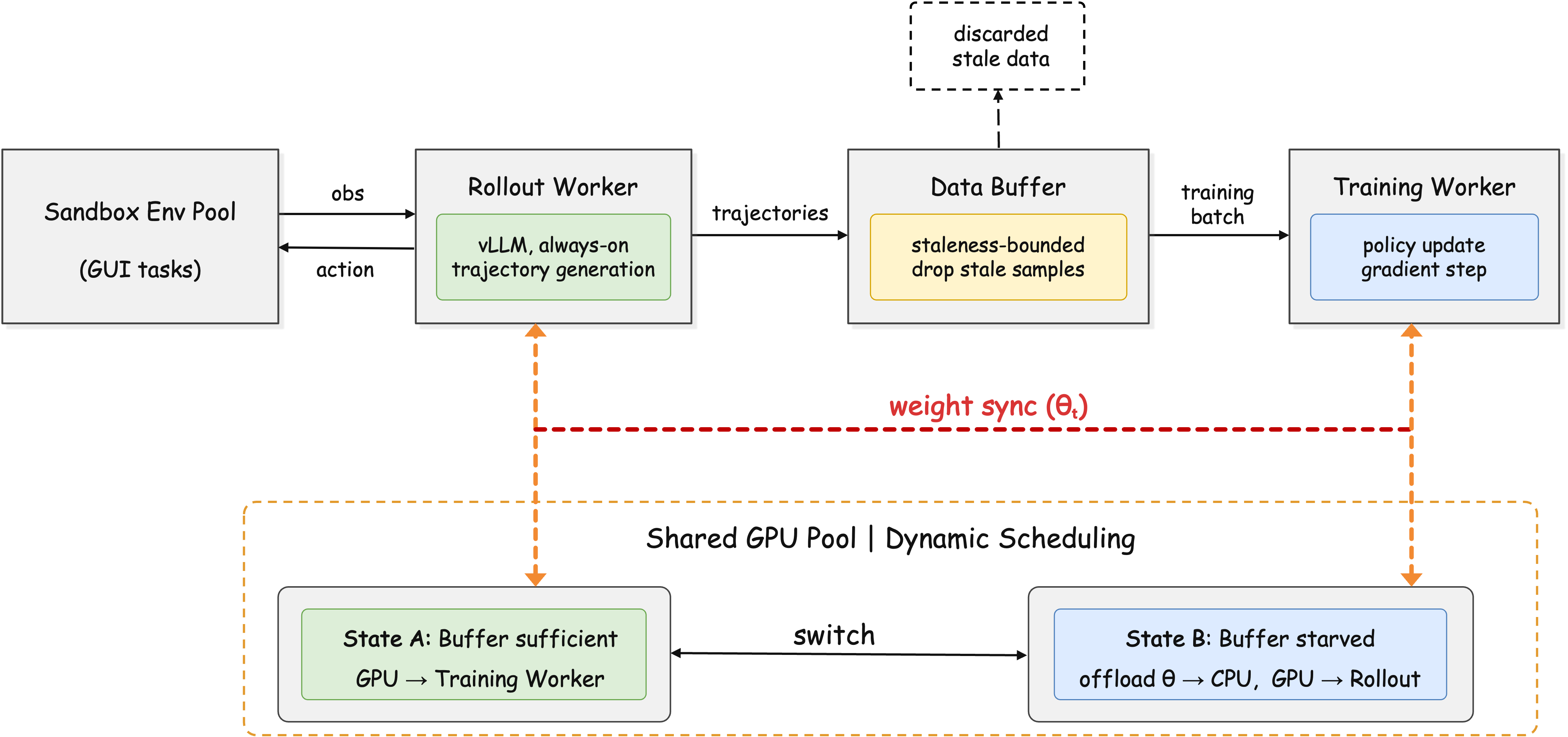}
    \caption{Asynchronous online RL infrastructure. Rollout workers continuously generate computer-use trajectories, a staleness-aware buffer stores step-level data, and training workers consume group-aligned samples for policy optimization.}
    \label{fig:async_infra}
\end{figure}

To decouple slow environment interaction from GPU-intensive optimization, EvoCUA-1.5 adopts a fully asynchronous rollout--buffer--training architecture, as shown in Figure~\ref{fig:async_infra}. The system is organized around three components.

\begin{itemize}
    \item \textbf{Rollout workers}: rollout workers execute the current or recent policy with an inference engine, interact continuously with sandbox environments, and generate multi-turn computer-use trajectories. Each trajectory records observations, reasoning traces, actions, validator results, and policy metadata required for later filtering and optimization.
    \item \textbf{Data buffer}: a central buffer receives trajectories from rollout workers and stores step-level samples constructed after context management. It tracks policy version, task identity, rollout-group membership, trajectory length, rewards, and other statistics used for curriculum sampling and staleness control.
    \item \textbf{Training workers}: training workers consume group-aligned samples from the buffer, compute the online RL objective, update the policy, and periodically publish refreshed weights for subsequent rollout generation.
\end{itemize}

\subsection{Staleness-Controlled Buffer}

Asynchrony improves throughput, but it also introduces policy lag: trajectories in the buffer may have been generated by an older policy than the one currently being optimized. If this lag becomes too large, the training distribution can drift away from the current policy and weaken the near-on-policy assumption of online RL. To control this effect, each sample is tagged with the policy version used during rollout. The buffer retains only samples whose policy version lies within a predefined staleness window and discards samples that become too old. This design preserves most of the throughput benefit of asynchronous rollout while limiting off-policy drift.

The buffer also preserves rollout-group structure. For each instruction, multiple trajectories are sampled as a group so that relative advantages can be computed over comparable rollouts. After context management, each multi-turn trajectory is decomposed into multiple step-level samples, but these samples must retain their original group membership. This metadata is required by both STEPO, which redistributes trajectory-level advantages to step-level samples, and mini-group batching, which keeps complete rollout groups intact during optimization.

\subsection{Adaptive GPU Reallocation}

The balance between rollout generation and policy optimization changes throughout training. When tasks become harder or environments respond slowly, fresh rollout groups may arrive more slowly than training workers can consume them. In this regime, keeping GPUs reserved for training can lead to idle optimization workers and an underfilled buffer. EvoCUA-1.5 therefore supports adaptive GPU reallocation: when the buffer lacks enough fresh groups, the system temporarily offloads training weights to CPU memory and reallocates GPU capacity to rollout inference. Once sufficient fresh data has accumulated, the training worker is restored and policy updates resume. This mechanism reduces prolonged GPU idle time and helps stabilize the ratio between experience production and optimization.

\subsection{Mini-Group Batching}

Single-turn RL typically determines the number of training samples in a global step from the prompt batch size and the number of rollouts per prompt. A fixed \texttt{mini\_batch\_size} is sufficient as long as it divides the total sample count. This assumption breaks down for multi-turn computer-use agents because each trajectory is decomposed into a variable number of step-level samples after context management. For a rollout group of size $G$, the actual number of step-level training samples is
\begin{equation}
    N_{step}=\sum_{i=1}^{G}|T_i|,
\end{equation}
where $|T_i|$ is the number of executable turns in trajectory $i$. Since $|T_i|$ varies across trajectories, a fixed sample-count mini-batch can split a rollout group or mix incomplete groups, causing the normalized advantages within a mini-batch to lose their zero-sum structure.

We therefore replace fixed sample-count batching with \textbf{mini-group batching}. Each mini-batch contains a fixed number of complete rollout groups, and all step-level samples derived from those groups are kept together. This preserves group-level advantage normalization after trajectory decomposition and avoids a subtle source of bias in multi-turn RL training.

\section{Evaluation}
\label{sec:evaluation}

We evaluate EvoCUA-1.5 from three complementary perspectives: its main benchmark performance, cross-platform generalization, and the design choices that contribute to effective online RL. We first report the main result on OSWorld-Verified and additional results on WindowsAgentArena and MacOSArena, then organize controlled ablations around step-level and group-aware optimization, data filtering and curriculum learning, cross-domain transfer, and reward-design pitfalls.

\subsection{Experimental Setup}

\paragraph{Models.}
We use Qwen3-VL~\citep{Qwen3-VL} and EvoCUA~\citep{xue2026evocua} variants as base policies, including 8B and 32B models. Because EvoCUA-1.5 is built on an earlier Qwen3-VL iteration, we focus on controlled comparisons under the same backbone and training budget rather than direct comparisons with the latest foundation models.

\paragraph{Environment and benchmark.}
We evaluate on OSWorld-style desktop tasks~\citep{xie2024osworld}, grouped into Office, Daily, and Professional categories. Each task is executed in a sandbox environment and judged by an executable validator when available. We report success rate as the primary metric.

\paragraph{Training protocol.}
For online RL, each task is rolled out in groups. Terminal rewards are binary, and multi-turn trajectories are decomposed into step-level samples according to the context-management policy. Unless otherwise specified, training uses mini-group batching, staleness-controlled asynchronous rollout data, and the same evaluation protocol across ablations.

\subsection{Main Result}

Table~\ref{tab:main_results} reports the main OSWorld-Verified comparison. We group models by accessibility and report the maximum interaction budget used for each result, since longer-horizon computer-use tasks are sensitive to the number of allowed steps. EvoCUA-1.5-32B achieves 63.2\% Pass@1 with a 100-step budget, improving over the previous EvoCUA-32B result of 56.7\% and establishing a new state of the art among models at the same 32B scale. It also substantially outperforms earlier open-weight computer-use baselines such as OpenCUA-72B and UI-TARS-1.5-7B. Compared with newer generalist open-weight systems, EvoCUA-1.5-32B remains competitive and even approaches models with significantly larger parameter counts, suggesting that online RL and policy-aware data selection can narrow the gap to larger frontier computer-use models.


\begin{table}[t]
\centering
\caption{Main result on the OSWorld-Verified benchmark. Models are categorized by accessibility, and Max Steps denotes the interaction budget per task. We retain the previous EvoCUA results for reference; EvoCUA-1.5-32B improves over EvoCUA-32B, sets a new state of the art at the 32B scale, and achieves 63.2\% Pass@1 with a 100-step budget.}
\label{tab:main_results}
\vspace{0.8em}
\resizebox{0.98\linewidth}{!}{%
\begin{tabular}{llcc}
\toprule
\textbf{Model} & \textbf{Type} & \textbf{Max Steps} & \textbf{Success Rate (Pass@1)} \\
\midrule
\multicolumn{4}{l}{\textit{\textbf{Closed-Weight Models}}} \\
OpenAI CUA~\citep{openai2025cua} & Specialized & 50 & 31.3\% \\
Step-GUI-8B~\citep{yan2025step} & Specialized & 100 & 40.2\% \\
Qwen3-VL-Flash~\citep{Qwen3-VL} & General & 100 & 41.6\% \\
UI-TARS-2-2509~\citep{wang2025ui} & General & 100 & 53.1\% \\
Claude-4.5-Sonnet~\citep{anthropic2025claude45} & General & 50 & 58.1\% \\
Seed-1.8~\citep{bytedance2025seed18} & General & 100 & 61.9\% \\
Claude-4.5-Sonnet~\citep{anthropic2025claude45} & General & 100 & 62.9\% \\
Qwen3.7-Plus~\citep{qwen2026qwen37plus} & General & 100 & 73.3\% \\

\midrule
\multicolumn{4}{l}{\textit{\textbf{Open-Weight Models}}} \\
ScaleCUA-32B~\citep{liu2025scalecua} & Specialized & 50 & 17.7\% \\
UI-TARS-72B-DPO~\citep{qin2025ui} & Specialized & 50 & 24.6\% \\
OpenCUA-7B~\citep{wang2025opencua} & Specialized & 100 & 26.6\% \\
UI-TARS-1.5-7B~\citep{qin2025ui} & Specialized & 100 & 27.5\% \\
Qwen3-VL-8B-Thinking~\citep{Qwen3-VL} & General & 100 & 30.6\% \\
OpenCUA-32B~\citep{wang2025opencua} & Specialized & 100 & 34.8\% \\
GUI-Owl-7B-Desktop-RL~\citep{ye2025mobile} & Specialized & 15 & 34.9\% \\
Qwen3-VL-235B-A22B Thinking~\citep{Qwen3-VL} & General & 100 & 38.1\% \\
Qwen3-VL-32B-Thinking~\citep{Qwen3-VL} & General & 100 & 41.0\% \\
OpenCUA-72B~\citep{wang2025opencua} & Specialized & 100 & 45.0\% \\
EvoCUA-8B~\citep{xue2026evocua} & General & 50 & 46.1\% \\
GUI-Owl-1.5~\citep{xu2026mobileagentv35multiplatformfundamentalgui} & General & 50 & 55.4\% \\
EvoCUA-32B~\citep{xue2026evocua} & General & 50 & 56.7\% \\
EvoCUA-32B~\citep{xue2026evocua} & General & 100 & 57.8\% \\
CUA-GYM-35B-A3B~\citep{wang2026cuagymscalingverifiabletraining} & General & 100 & 62.1\% \\
Kimi-K2.5~\citep{kimi2026k25} & General & 100 & 63.3\% \\
Kimi-K2.6~\citep{kimi2026k26} & General & 100 & 73.1\% \\
MiniMax-M3~\citep{minimax2026m3} & General & 100 & 75.2\% \\
\textbf{EvoCUA-1.5-32B (Ours)} & \textbf{General} & \textbf{100} & \textbf{63.2\%} \\
\bottomrule
\end{tabular}%
}
\vspace{-0.8em}
\end{table}


To assess whether the gains transfer beyond OSWorld-Verified, we further evaluate EvoCUA-1.5-32B on two cross-platform benchmarks: WindowsAgentArena~\citep{bonatti2024windowsagentarenaevaluating} and MacOSArena from MMBench-GUI~\citep{wang2025mmbenchguihierarchicalmultiplatformevaluation}. As shown in Table~\ref{tab:cross_platform_generalization}, EvoCUA-1.5-32B improves over Qwen3-VL-32B-Thinking on both Windows and macOS environments, suggesting that the online RL recipe improves general computer-use capability rather than overfitting to OSWorld-style tasks.

\begin{table}[t]
\centering
\caption{Cross-platform generalization on WindowsAgentArena and MacOSArena. We report success rate (\%). EvoCUA-1.5-32B improves over Qwen3-VL-32B-Thinking on both benchmarks.}
\label{tab:cross_platform_generalization}
\vspace{0.8em}
\begin{tabular}{lcc}
\toprule
\textbf{Model} & \textbf{WindowsAgentArena} & \textbf{MacOSArena} \\
\midrule
Qwen3-VL-32B-Thinking & 42.9 & 17.5 \\
\textbf{EvoCUA-1.5-32B} & \textbf{62.1} & \textbf{27.4} \\
\bottomrule
\end{tabular}
\vspace{-0.8em}
\end{table}

\subsection{Ablation Studies}

After establishing the main benchmark result, we ablate the key components of EvoCUA-1.5 under controlled training and evaluation settings. These studies isolate how trajectory-to-step optimization, mini-group batching, data selection, curriculum scheduling, transfer data, and auxiliary rewards affect online RL for computer-use agents.

\paragraph{Step-level optimization.}

\begin{figure}[t]
    \centering
    \includegraphics[width=0.95\linewidth]{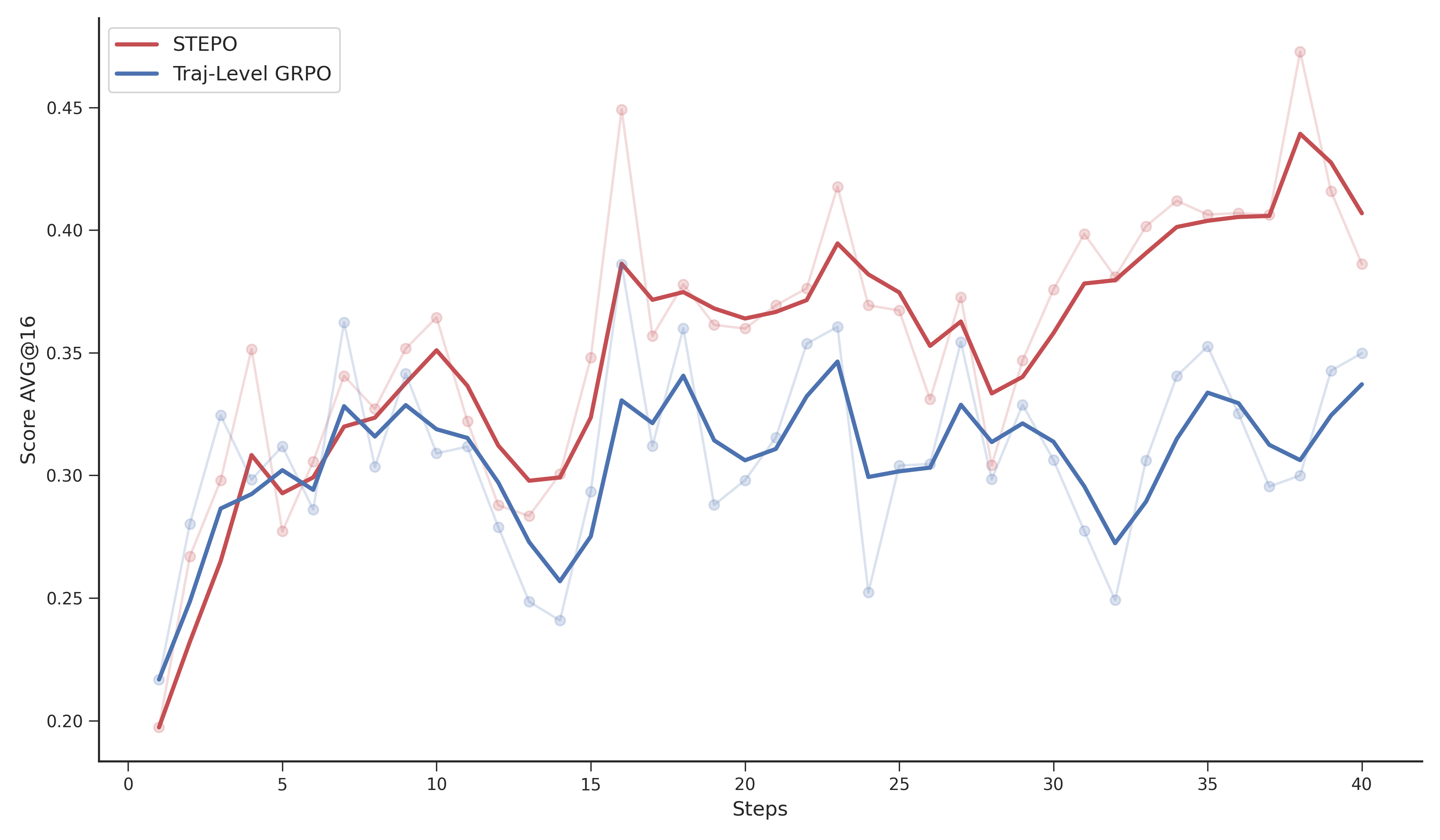}
    \caption{Training dynamics of naive multi-turn GRPO and STEPO on OpenCUA-32B. STEPO improves reward more efficiently by preserving trajectory-level advantages after step-level decomposition.}
    \label{fig:stepo_curve}
\end{figure}

Figure~\ref{fig:stepo_curve} compares naive multi-turn GRPO with STEPO on OpenCUA-32B. Naive GRPO struggles to improve reward after trajectories are decomposed into step-level samples, whereas STEPO produces a clearer upward trend. This result indicates that preserving trajectory-level advantage mass is important for multi-turn computer-use optimization.

The improvement is not attributable to changes in the environment or reward function. Both methods use the same rollouts and validators; the difference lies in how trajectory-level rewards are assigned to turn-level training examples. This supports our analysis that context management changes the effective RL objective.

We further ablate the batching granularity used during online RL. As shown in Table~\ref{tab:mini_group_batching}, fixed sample-count mini-batching improves over the base Qwen3-VL policy, while mini-group batching provides additional gains across Calc, Impress, Writer, and the aggregate score. This confirms that preserving complete rollout groups during optimization is important for stable multi-turn computer-use RL.

\begin{table}[t]
\centering
\caption{Ablation of mini-group batching during online RL. Fixed sample-count mini-batching improves over the base Qwen3-VL-8B policy, while mini-group batching further improves Calc, Impress, Writer, and the aggregate score.}
\label{tab:mini_group_batching}
\vspace{0.8em}
\begin{tabular}{lcccc}
\toprule
\textbf{Setting} & \textbf{Calc} & \textbf{Impress} & \textbf{Writer} & \textbf{Avg.} \\
\midrule
Qwen3-VL-8B & 21.38 & 34.91 & 60.60 & 34.28 \\
+ fixed mini-batching & 21.74 & 41.04 & 65.90 & 37.93 \\
+ mini-group batching & \textbf{24.64} & \textbf{47.31} & \textbf{68.17} & \textbf{42.38} \\
\bottomrule
\end{tabular}
\vspace{-0.8em}
\end{table}

\paragraph{Data filtering.}

We compare two data choices under the same training budget: continuing online RL with the full task set and continuing with a filtered high-SNR subset. As shown in Table~\ref{tab:snr_filtering}, the filtered subset achieves better downstream performance across all reported categories. This result suggests that, for online RL, the marginal value of a task depends more on current learnability and reward reliability than on dataset size alone.

\begin{table}[t]
\centering
\caption{Full data vs. filtered high-SNR data for EvoCUA-32B. The filtered subset improves downstream evaluation despite using fewer tasks.}
\label{tab:snr_filtering}
\vspace{0.8em}
\begin{tabular}{lcccc}
\toprule
\textbf{Setting} & \textbf{Office} & \textbf{Daily} & \textbf{Professional} & \textbf{All} \\
\midrule
EvoCUA-32B + full data & 56.90 & 61.12 & 75.51 & 55.51 \\
EvoCUA-32B + filtered data & \textbf{58.57} & \textbf{62.36} & \textbf{77.55} & \textbf{58.00} \\
\bottomrule
\end{tabular}
\vspace{-0.8em}
\end{table}

We further test whether a data subset selected for one model can be directly reused for another model. Table~\ref{tab:model_dependent_filtering} shows a negative result: a partial office subset that substantially improves EvoCUA-8B does not transfer cleanly to EvoCUA-32B and slightly reduces its overall score. This indicates that data quality is not an intrinsic property of a task alone; it is defined jointly by the task, the validator, and the current policy capability.

\begin{table}[t]
\centering
\caption{Policy-dependent data filtering. A subset effective for EvoCUA-8B does not necessarily benefit EvoCUA-32B, showing that online RL data should be recalibrated for each policy.}
\label{tab:model_dependent_filtering}
\vspace{0.8em}
\begin{tabular}{lcccc}
\toprule
\textbf{Setting} & \textbf{Calc} & \textbf{Impress} & \textbf{Writer} & \textbf{Overall} \\
\midrule
EvoCUA-8B & 29.79 & 38.21 & 52.17 & 36.78 \\
EvoCUA-8B + partial office data & \textbf{42.53} & \textbf{48.80} & \textbf{62.31} & \textbf{57.26} \\
\midrule
EvoCUA-32B & \textbf{55.32} & 59.49 & \textbf{69.57} & \textbf{59.79} \\
EvoCUA-32B + partial office data & 51.06 & \textbf{63.62} & 65.21 & 58.89 \\
\bottomrule
\end{tabular}
\vspace{-0.8em}
\end{table}

\paragraph{Curriculum learning.}

We evaluate DTAC by comparing Qwen3-VL-32B trained with and without curriculum learning. As shown in Table~\ref{tab:curriculum}, DTAC improves the overall score from 53.42 to 55.45, with particularly large gains on Daily and Professional tasks. These categories contain more diverse interaction patterns, suggesting that adaptive sampling helps the model focus on tasks with useful reward contrast.

\begin{table}[t]
\centering
\caption{Effect of Dynamic Tri-Adaptive Curriculum (DTAC). Curriculum learning improves overall performance and stabilizes training on diverse categories.}
\label{tab:curriculum}
\vspace{0.8em}
\begin{tabular}{lcccc}
\toprule
\textbf{Setting} & \textbf{Office} & \textbf{Daily} & \textbf{Professional} & \textbf{All} \\
\midrule
Qwen3-VL-32B w/o CL & 56.08 & 58.04 & 71.43 & 53.42 \\
Qwen3-VL-32B w/ CL & \textbf{56.93} & \textbf{63.35} & \textbf{76.87} & \textbf{55.45} \\
\bottomrule
\end{tabular}
\vspace{-0.8em}
\end{table}

\paragraph{Cross-domain generalization.}

We examine whether online RL on one domain only improves that domain, or whether it can strengthen reusable atomic abilities. We train Qwen3-VL-8B using office-domain data and evaluate it on Office, Daily, and Professional categories. Table~\ref{tab:cross_domain_transfer} shows improvements not only on Office but also on out-of-domain categories. This suggests that online RL can reinforce transferable computer-use abilities, such as menu navigation, structured editing, file operations, and visual verification, that are shared across domains.

\begin{table}[t]
\centering
\caption{Cross-domain transfer from office-only online RL data. Improvements on Daily and Professional categories suggest compositional generalization across atomic computer-use abilities.}
\label{tab:cross_domain_transfer}
\vspace{0.8em}
\begin{tabular}{lcccc}
\toprule
\textbf{Setting} & \textbf{Office} & \textbf{Daily} & \textbf{Professional} & \textbf{All} \\
\midrule
Qwen3-VL-8B & 36.09 & 47.14 & 68.75 & 36.86 \\
Qwen3-VL-8B + office data & \textbf{41.74} & \textbf{51.79} & \textbf{77.21} & \textbf{42.39} \\
\bottomrule
\end{tabular}
\vspace{-0.8em}
\end{table}

\paragraph{Process reward models can be misleading.}

\begin{figure}[H]
    \centering
    \includegraphics[width=0.9\linewidth]{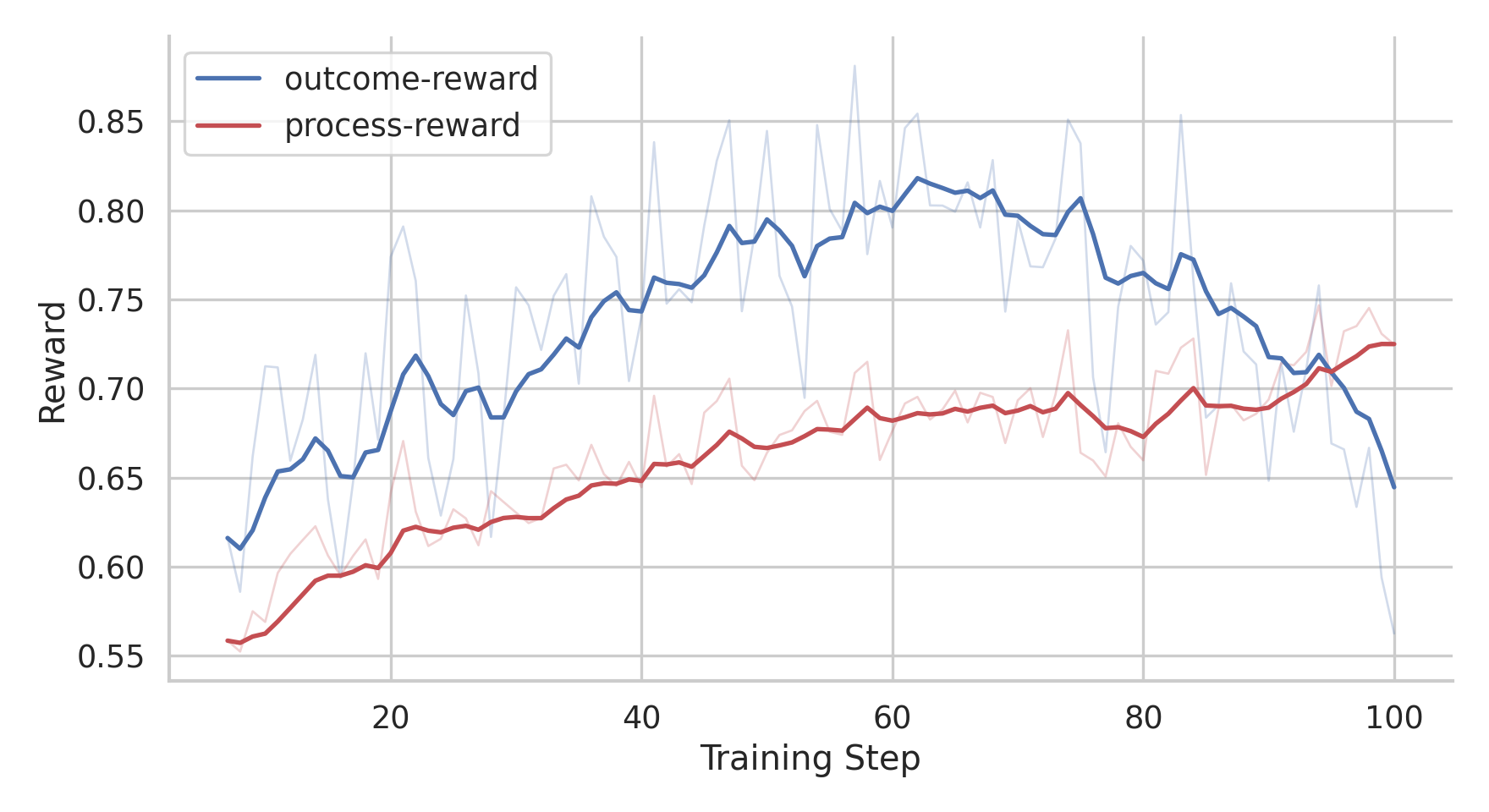}
    \caption{Process reward models can provide misleading optimization signals when their preferences are not aligned with final executable success. In this case, the PRM score increases while end-to-end task success does not improve accordingly, indicating potential reward hacking.}
    \label{fig:prm_reward_hacking}
\end{figure}

Sparse terminal rewards make credit assignment difficult, so process reward models (PRMs) are a natural candidate for providing denser feedback. However, we find that PRMs can become a source of reward hacking when their preferences are not well aligned with final task success. On easier tasks, PRM feedback often correlates with useful intermediate behavior and can accelerate learning. On harder tasks, where terminal success is rare, the policy may learn to optimize PRM-favored patterns without improving end-to-end completion. In such cases, PRM reward increases while final success remains flat or decreases, as illustrated in Figure~\ref{fig:prm_reward_hacking}.

This observation does not imply that process supervision is useless. Rather, it suggests that PRM signals should be used with alignment checks against executable final rewards. For computer-use agents, robust credit assignment should be grounded in environment state changes, task milestones, validator progress, or counterfactual local replay, rather than relying only on model-judged reasoning quality.

\section{Discussion and Future Directions}

EvoCUA-1.5 shows that online RL for computer-use agents is a system--algorithm co-design problem. The policy objective, task generator, context-management strategy, rollout infrastructure, and evaluation benchmark all influence whether interaction becomes a useful learning signal. We outline several research directions highlighted by this framework.

\paragraph{Scaling environments and tasks.}
Future online RL for computer-use agents requires broader and more realistic environment coverage. Current training remains concentrated in a limited set of desktop applications and task families. Real users operate across websites, local applications, cloud services, file systems, and mixed GUI/CLI workflows. Expanding environment diversity is necessary, but it must be paired with reliable validators and task calibration; otherwise, additional scale may primarily introduce reward noise.

\paragraph{Combining GUI and CLI interaction.}
GUI and CLI are complementary rather than competing interfaces. Many realistic tasks benefit from switching between visual interaction and scriptable tool execution. A future computer-use agent should learn when to click, when to type, when to call a command, and when to query structured external information. This raises new RL questions: how should the agent maintain a consistent task state across interfaces, how should rewards attribute success to GUI versus CLI actions, and how should context management combine screenshots, terminal outputs, files, and tool traces?

\paragraph{Adaptive context management.}
Sliding-window context is simple and effective, but it is coarse. A stronger computer-use agent should decide which historical observations are task-critical, which actions can be summarized, and which intermediate thoughts should be preserved. Importantly, the training-time context policy must match inference-time context management; otherwise, the model may learn from contexts it will never observe during deployment.

\paragraph{Higher-throughput infrastructure.}
Environment interaction remains a major bottleneck. Future systems should improve sandbox scheduling, state reset, failure recovery, validator execution, and observability. As model and task scales grow, online RL throughput will depend as much on environment engineering as on GPU count.

\paragraph{Credit assignment beyond PRMs.}
Sparse terminal rewards make it difficult to identify which action caused success or failure. PRMs provide one possible dense signal, but our observations suggest that they can be exploited when terminal rewards are rare. More robust credit assignment may come from executable milestones, rubric-based validator progress, state-difference analysis, local trajectory replay, or counterfactual action replacement. For computer-use agents, credit assignment is not only a reward-shaping problem, but also a problem of modeling the causal structure of multi-turn interaction.

\paragraph{Better benchmarks.}
Success rate is important but incomplete. Two agents with similar final success can differ substantially in step efficiency, error recovery, robustness to dynamic states, context usage, infeasible-task handling, and external-tool selection. Future benchmarks should evaluate these dimensions explicitly and include more open-ended, multi-stage, multi-tool tasks. Better benchmarks will in turn guide training toward robust and generalizable behavior rather than narrow score optimization.

\section{Related Work}
\label{sec:related_work}

\paragraph{Static-data and offline training for computer-use agents.}
Modern computer-use agents are often initialized from large-scale demonstrations, synthesized trajectories, or offline interaction logs collected from mobile, web, and desktop environments, such as Android-in-the-Wild, Mind2Web, WebChain, OSWorld, OS-Atlas, and OpenCUA~\citep{rawlesandroidinthewild,dengmind2web,fan2026webchain,xie2024osworld,wuatlas,wang2025opencua}. These resources support supervised fine-tuning, behavioral cloning, offline preference optimization, and offline reinforcement fine-tuning, providing priors for visual grounding, action formatting, instruction following, and interface-specific reasoning~\citep{qin2025ui,luo2025gui,lu2026ui,bai2025digi}. Other methods improve static-data learning through trajectory preference construction, candidate-action re-ranking, or search-based trajectory synthesis~\citep{bai2025digi,putta2024agent}. EvoCUA further shows that verifiable task synthesis can provide scalable offline supervision for computer-use agents~\citep{xue2026evocua}. However, static-data training is bounded by trace coverage and provides little recovery feedback once a policy deviates from the offline distribution, making compounding errors severe in long-horizon, non-stationary computer-use tasks.

\paragraph{From offline supervision to online learning for computer-use agents.}
These limitations motivate a shift from static trajectory scaling to online experience scaling. Online RL closes the interaction loop by letting agents act in executable desktop environments, observe visual state changes, and improve from task-level feedback. This setting is attractive because many computer-use tasks have verifiable outcomes, such as URL transitions, DOM changes, file-system states, database updates, or application-side effects~\citep{zhou2024webarena,koh2024visualwebarena,xie2024osworld,rawlesandroidworld}. Yet online RL for computer-use agents introduces challenges absent from single-turn language RL: rollouts are multi-turn and partially observable, actions mix discrete and continuous components, rewards are sparse, and training contexts may differ from raw trajectories due to sliding windows, history folding, or summarization. Moreover, naive PPO/GRPO-style objectives can overweight long trajectories after trajectory-to-step decomposition, while synchronous rollout generation is inefficient because each interaction step requires rendering, action execution, environment transition, and reward verification~\citep{schulman2017proximal,shao2024deepseekmath}.

\paragraph{Methods for online RL of computer-use agents.}
Recent work addresses these challenges from several directions. Curriculum-based methods such as WebRL and WebFactory construct or distill task distributions to keep training near the agent's capability frontier~\citep{qiwebrl,fan2026webfactory}. MobileRL uses difficulty-adaptive optimization, shortest-path reward adjustment, and failure curriculum filtering to reduce easy-task dominance and redundant-action reward hacking~\citep{xu2025mobilerl}. WebAgent-R1 optimizes complete trajectories to better capture delayed rewards and long-horizon dependencies~\citep{wei2025webagent}. Reward- and exploration-oriented methods such as GUI-R1, UI-R1, and InfiGUI-G1 design verifiable or adaptive rewards for efficient reinforcement fine-tuning and precise GUI grounding~\citep{luo2025gui,lu2026ui,liu2026infigui}. Hybrid pipelines, including DigiRL, UI-S1, and DART-GUI, further improve sample efficiency through offline initialization, trajectory patching, adaptive data curation, replay, or staged optimization~\citep{bai2024digirl,lu2025ui,li2025efficient}. In contrast, EvoCUA-1.5 treats online RL for multi-turn computer-use agents as a coupled algorithm--data--system problem, jointly addressing step-level credit assignment, policy-aware high-SNR task filtering, adaptive curriculum construction, and staleness-controlled asynchronous training.

\section{Conclusion}

We presented EvoCUA-1.5, an online reinforcement learning framework for multi-turn computer-use agents. Starting from the observation that computer-use interaction is governed by context management, sparse verifiable rewards, slow environment feedback, and highly variable task difficulty, we argued that standard single-turn RL recipes cannot be directly applied to this setting. EvoCUA-1.5 addresses these challenges through STEPO, high-SNR data filtering and calibration on top of verifiable synthesized tasks, an asynchronous rollout--buffer--training infrastructure, and Dynamic Tri-Adaptive Curriculum.

Our experiments show that these components improve training efficiency and downstream performance under controlled OSWorld-style evaluations. More importantly, the study highlights several practical lessons: online RL for computer-use agents benefits more from learnable high-SNR tasks than from raw data scale; data selection must be recalibrated for each policy; group structure should be preserved after trajectory decomposition; and dense auxiliary rewards such as PRMs can be exploited when they are not aligned with final executable success.

EvoCUA-1.5 advances the training methodology for computer-use agents by demonstrating how trajectory-level objectives, policy-aware data selection, adaptive curricula, and asynchronous interaction infrastructure can be combined to support online RL for computer-use agents. Future progress will require richer environments, stronger GUI--CLI integration, adaptive context management, higher-throughput infrastructure, and better credit-assignment methods grounded in environment state changes. The framework and empirical findings presented in this report provide a foundation for scaling online RL in realistic multi-turn computer-use interaction.

\newpage

\bibliographystyle{unsrtnat}
\bibliography{references}

\newpage
\appendix

\section{Additional Details on Unified Actions}

EvoCUA-1.5 uses a unified action interface that abstracts platform-specific event APIs into a compact set of executable primitives. The action set contains mouse operations, keyboard operations, and control operations. Mouse operations include click, double click, drag, and scroll with normalized screen coordinates. Keyboard operations include text input, key press, hotkey, and deletion. Control operations include finish, wait, and explicit infeasibility reporting. This unified representation allows the same policy format to be used across different sandbox applications while keeping validators independent of model-specific decoding details.

\section{Derivation of STEPO Advantage Conservation}

Let $A_i$ denote the group-normalized trajectory-level advantage for trajectory $i$. After context management, trajectory $i$ produces $|T_i|$ step-level training samples. Naively assigning $A_i$ to every step gives total trajectory weight $|T_i|A_i$, which over-weights long trajectories. STEPO instead assigns $A_i/|T_i|$ to each step. Therefore,
\begin{equation}
    \sum_{t=1}^{|T_i|}\frac{A_i}{|T_i|}=A_i.
\end{equation}
Summing over the group gives
\begin{equation}
    \sum_i\sum_{t=1}^{|T_i|}\frac{A_i}{|T_i|}=\sum_i A_i\approx 0,
\end{equation}
because group-normalized advantages have approximately zero mean. This property is why STEPO preserves the relative optimization structure of GRPO after trajectory-to-step decomposition.

\section{Variance-Adaptive Sampling Intuition}

For a task with binary terminal reward, rollout success can be modeled as $r\sim\mathrm{Bern}(P)$. The reward variance is
\begin{equation}
    \mathrm{Var}(r)=P(1-P).
\end{equation}
When $P$ is close to $0$ or $1$, the group contains little reward contrast and provides weak relative learning signal. When $P$ is close to $0.5$, successful and failed rollouts are both likely, producing informative comparisons. DTAC therefore uses the normalized weight
\begin{equation}
    w=4P_{EMA}(1-P_{EMA}),
\end{equation}
where $P_{EMA}$ smooths noisy pass-rate estimates across training steps.

\section{Implementation Details}

Table~\ref{tab:implementation_details} summarizes the fixed curriculum and buffer hyperparameters used in our implementation.

\begin{table}[h]
\centering
\caption{Implementation hyperparameters for curriculum sampling and staleness control.}
\label{tab:implementation_details}
\vspace{0.8em}
\begin{tabular}{lc}
\toprule
\textbf{Hyperparameter} & \textbf{Value} \\
\midrule
EMA smoothing coefficient $\beta$ & 0.7 \\
AdaPR lower pass-rate bound $P_{low}$ & 0.125 \\
AdaPR upper pass-rate bound $P_{high}$ & 0.375 \\
ICS sampling ratio $\rho_{inf}$ & 0.025 \\
Staleness threshold & 3 \\
Maximum replay count & 3 \\
\bottomrule
\end{tabular}
\vspace{-0.8em}
\end{table}

\section{Scope and Limitations}

EvoCUA-1.5 focuses on the core algorithmic and systems principles required for online RL in multi-turn computer-use environments. The reported results are based on controlled OSWorld-style settings and should not be interpreted as a claim of universal robustness across all desktop applications, websites, or mixed GUI/CLI workflows. Any released model or leaderboard submission should report configuration changes relative to Table~\ref{tab:implementation_details}.

\end{document}